# Vulgar Remarks Detection in Chittagonian Dialect of Bangla

Tanjim Mahmud[1], Michal Ptaszynski[1], Fumito Masui[1]

[1]Text Information Processing Laboratory, Kitami Institute of Technology, Kitami, Japan
tanjim_cse@yahoo.com, {michal,f-masui}@mail.kitami-it.ac.jp

**Abstract**
The negative effects of online bullying and harassment are increasing with Internet popularity, especially in social media. One solution is using natural language processing (NLP) and machine learning (ML) methods for the automatic detection of harmful remarks, but these methods are limited in low-resource languages like the Chittagonian dialect of Bangla.This study focuses on detecting vulgar remarks in social media using supervised ML and deep learning algorithms.Logistic Regression achieved promising accuracy (0.91) while simple RNN with Word2vec and fastTex had lower accuracy (0.84-0.90), highlighting the issue that NN algorithms require more data.

## 1. Introduction

The use of vulgar language, swearing, taboo words or other offensive language is referred to as vulgarity or obscenity (Cachola et al., 2018; Wang, 2013). Although in society, vulgar language in conversation is frequent (Mehl et al., 2007), it has become very popular on social media sites like Twitter (Wang et al., 2014). Although vulgar language can be employed in positive context, such as indicating informality of dialogue, communicating one's anger, or identifying with a group (Holgate et al., 2018), in reality it is most often used in online harassment. Therefore, in our study we focused on detecting such vulgar expressions in Chittagonian dialect of Bangla. Chittagonian is a language from the Indo-Aryan language family [1] with between 13 and 16 million speakers, the great majority of them living in Bangladesh (LewisM, 2009). Chittagonian is widely spoken alongside Bengali, to the extent that many linguists consider it a distinct language (Masica, 1993).It is a variation of Bangla, with distinct features in pronunciation, vocabulary, and grammar.

Over the last two decades, Internet use in Bangladesh has grown exponentially. According to BTRC, there are well over 125 million Internet users in Bangladesh as of November 2022[2]. In addition, Chittagong is the second largest city of Bangladesh[3], and due to the Digital Bangladesh initiative[4], the majority of the population now has access to the Internet and is able to use social media. Moreover, with the benefit of Unicode on gadgets, they express their thoughts in their native Chittagonian dialect. Chittagonian people regularly use such social media as Facebook[5], imo[6], various blogs, WhatsApp[7], and others. On social networking sites, people feel free to express themselves in casual ways. However, due to its wide pervasiveness, it is difficult to escape also the negative influence of social media. Excessive social media use has the potential to become addictive[8]. Young people spend more time on social media than they do with family and friends[9]. Social media use has been linked to cyberbullying and online abuse, which has an impact on self-esteem and can be a violation of one's privacy[10]. Social media has also been used to disseminate hatred and deception online, leading to an increase in violent incidents in society[11]. One of the realizations of such unwanted and harmful incidents is receiving messages filled with vulgar expressions. Moreover, with the increased usage of social media comes the increased probability of being exposed to such vulgar remarks.

To contribute to the mitigation of this problem, in this work, we propose a system capable of automatically recognizing such vulgar remarks. Using Logistic Regression (LR) and Recurrent Neural Networks (RNN) as a classifier, backed by various feature extraction methods, we test the limitations of such methods for vulgar remark detection in the low-resource language scenario.

The key contributions of this paper are as follows:

1. We collect a dataset in the Chittagonian dialect consisting of 2,500 comments or posts. The data was gathered purely from widely accessible accounts on the Facebook platform.
2. We manually and rigorously annotate the dataset into vulgar and non-vulgar categories and validate

---

[1]https://en.wikipedia.org/wiki/Chittagonian_language
[2]http://www.btrc.gov.bd/site/page/347df7fe-409f-451e-a415-65b109a207f5/-
[3]https://en.wikipedia.org/wiki/Chittagong
[4]https://www.undp.org/bangladesh/blog/digital-bangladesh-innovative-bangladesh-road-2041
[5]https://www.facebook.com
[6]https://imo.im
[7]https://www.whatsapp.com
[8]https://www.addictioncenter.com/drugs/social-media-addiction/
[9]https://en.prothomalo.com/bangladesh/Youth-spend-80-mins-a-day-in-internet-adda
[10]https://www.un.org/en/chronicle/article/cyberbullying-and-its-implications-human-rights
[11]https://www.accord.org.za/conflict-trends/social-media/

the data annotation process by Cohen's Kappa statistics (Cohen, 1960).
3. Finally, we compare Machine Learning (ML)-based, and Deep Learning(DL)-based approaches for identifying vulgar remarks in the Chittagonian dialect on social media content.

The paper is organized as follows: Section 2. covers related works. Section 3. presents the proposed technique, including data collection and pre-processing. Section 4. discusses evaluation and analysis. Finally, Section 5. offers concluding remarks and future work.

## 2. Related works

Vulgar language in user-generated content on social media can lead to sexism, racism, hate speech, or other forms of online abuse (Cachola et al., 2018). Vulgarity detection, while typically solved by creating lexicons of vulgar expressions, can also be treated as a classification problem with two classes: vulgar and not. Traditional methods using vulgarity lexicons require constant updates. ML methods can use the surrounding context of vulgarities to classify new vulgarities without a lexicon. Few studies have approached vulgarity detection with methods beyond lexicon-based, so in this review, we also included studies from closely related domains. Many linguistic and psychological studies have been conducted on the purposes and pragmatic goals of vulgar language (Andersson and Trudgill, 1990; Pinker, 2007; Wang, 2013). On the other hand for the machine learning-related studies, for example, (Eshan and Hasan, 2017) compared various ML algorithms with N-gram features to find out which algorithms perform better. With 2,500 comments they labeled their dataset into two classes. Finally, by SVM with trigram TF-IDF vectorizer features they got the highest accuracy of 0.89. (Akhter et al., 2018) suggested using machine learning methods and using user data to identify cyberbullying in the Bangla language. They applied NB, J48, SVM, and KNN, with the performance of each method evaluated using a 10-fold cross-validation. According to the results, SVM performed better in terms of Bangla text, with the highest accuracy of 0.9727. (Holgate et al., 2018) introduced a dataset of 7,800 tweets from users with known demographics, every incident containing vulgarities was categorized into one of the six categories of vulgar word use. They examined the pragmatic components of vulgarity and their relationships to societal issues using the data they collected, obtaining 0.674 of macro F1 over six classes. (Emon et al., 2019) created a system for detecting abusive Bengali text and applied different deep learning and machine learning-based algorithms. They collected 4,700 comments from Facebook, Youtube, Prothom Alo online and labeled this data into seven different classes. They got the highest accuracy of 0.82 with the Recurrent Neural Network algorithm. (Awal et al., 2018) proposed Naive Bayes system to detect abusive comments and collected 2,665 English comments from Youtube and then translated these English comments into Bengali in two ways, namely, i) Direct translation to

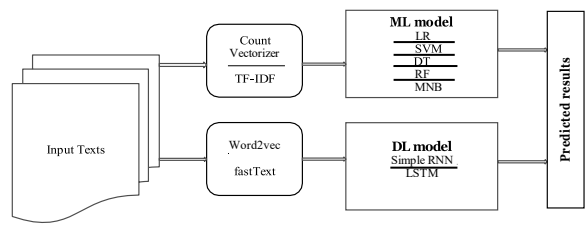

Figure 1: Outline of performed experiments.

Bangla, and ii) Dictionary-based translation to Bangla. Finally, their proposed system produced the highest accuracy of 0.8057.

For detecting abusive Bangla comments a technique developed by (Hussain and Al Mahmud, 2019) applied a root-level algorithm with unigram string features. They collected 300 comments from Facebook pages, news portals, and Youtube. They divided their data set into three sets with 100, 200, 300 comments and tested their system with the result of 0.689 (accuracy) on average. (Das et al., 2022) studied hate speech detection in low resourced language Bengali and Romanized Bengali. They collected their data from Twitter which contain 5071 Bengali samples, and 5107 Romanized Bengali samples. For training these datasets they used XML-RoBERTa, MuRIL, m-BERT and IndicBERT models, from which XML-RoBERTa gives the best accuracy of 0.796. (Sazzed, 2021) manually separated 7,245 YouTube reviews into categories of vulgar and non-vulgar content to establish two benchmark corpora. The DL-based BiLSTM model produced the highest recall scores for detecting vulgarity in both datasets. (Faisal Ahmed et al., 2021) marked user comments from publicly visible Facebook postings made by sportsmen, public servants, and celebrities. Then, the English- and mixed-language comments were separated from the Bengali-language comments. Their research showed that 14,051 of all remarks, or 31.9% of the total, were directed at male victims, and 29,950 of all, or 68.1%, were directed at female victims. In this study, 9,375 comments were directed towards victims who were social influencers, followed by 2,633 comments directed at politicians, 2,061 comments directed at athletes, 2,981 comments directed at singers, and 61.25% of the remarks, or 26,951, directed at actors.

None of the preceding research particularly looked for vulgarities in the Chittagong dialect of Bengali. In the context of the data from the Chittagong dialect on social media, the following is the first effort to precisely identify and evaluate the frequency of vulgarity in social media posts.

## 3. Proposed methodology

The layout of the proposed system is demonstrated in Figure 1 and the process is explained as follows.

### 3.1. Data collection

Since there is no state-of-the-art dataset for Chittagonian dialect to detect vulgar texts, one of the first major

| English | Translation |
|---|---|
| অডা তুই শুয়োরের বাচ্চা | You are piglet |
| তোরা বেগুন খানকির ফুয়া | You are all son of whores |
| বাংলাদেশত নতুন বেইশ্যাদেহা যারা | New prostitutes are appearing in Bangladesh. |

Table 1: Three examples from dataset.

| Chittagonian Dialect | English | Frequency |
|---|---|---|
| মাগির | Slut's | 333 |
| চোদা | Fucker | 201 |
| কুত্তা | Dog | 192 |
| খানকি | Whore | 105 |
| সোনা | Female private parts | 75 |

Table 2: Top frequency vulgar words in our dataset.

Original Text
ছুদানির ফুয়ার গলাধে গলা।1 No.(...)
English Translation(Literally)
The son of a slut's singing voice is very sweet.1 number(...)
Removing Functuations
ছুদানির ফুয়ার গলাধে গলা 1 No(...)
Removing Emojis and Emoticons
ছুদানির ফুয়ার গলাধে গলা 1 No
Removing English Characters
ছুদানির ফুয়ার গলাধে গলা 1
Removing English Digits
ছুদানির ফুয়ার গলাধে গলা
Removing Stopwords
ছুদানির ফুয়ার গলাধে গলা
Tokenizations
['ছুদানির',' ফুয়ার',' গলাধে',' গলা']

Table 3: Step-by-step data preprocessing.

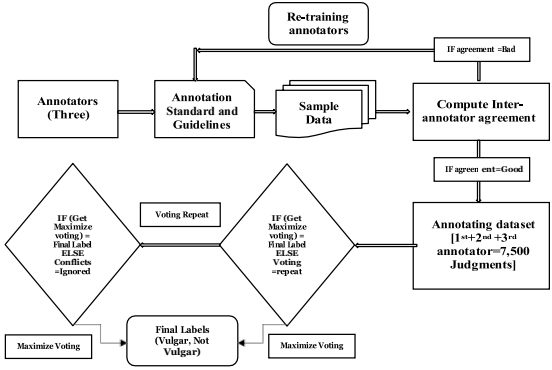

Figure 2: Data annotation process.

challenges in this study was collecting the data. We collected comments from Facebook. The comments were collected manually from several independent sources including public profiles and pages of famous people. Table 1 shows three examples from the dataset.

### 3.2. Data annotation
To detect vulgarity, a dataset needed to be annotated in accordance with a set of standards (Pradhan et al., 2020; Khan et al., 2021). We appointed three native Chittagonian dialect speakers, one of whom was a male with a higher education (MSc degree) and two of whom were women (BSc degree). The dataset contained 2,500 samples and each of them was manually annotated according to the process shown in Figure 2. Firstly, three annotators annotated each review which resulted in 7,500 judgments. Any disagreements about the annotation were resolved by majority voting of the annotators. Consequently, the raw dataset consisted of 1,009 vulgar samples, 1,476 non-vulgar samples with 15 conflicts (discarded from the dataset after discussion with annotators). Some of the most frequently used vulgar words in our dataset were shown in Table 2. After annotating the data, we examined the inter-rater agreement. As a result, using Cohen's Kappa (Cohen, 1960), we obtained an average agreement value of 0.91, indicating very strong agreement among annotators.

### 3.3. Data preprocessing
Our collected dataset of text-based comments was not in fixed length, and does not follow any specific structure, which means it could contain noise. The dataset comments could contain excessive data unimportant for analysis. To limit the possible influence of such unwanted redundant features, we processed the data. Table 3 shows each preprocessing step.

### 3.4. Feature engineering
Since ML algorithms cannot take as input textual data directly, it is necessary to convert lexical features (words) into numerical features in order to be able to extract patterns and perform classification. Since after preprocessing our dataset contains only strings of characters (words), we transformed the textual data into numerical features. To do this, following previous studies (Emon et al., 2019; Sazzed, 2021; Mahmud et al., 2022) we applied four different feature extraction techniques, namely, Count Vectorizer, TF-IDF Vectorizer, Word2vec and fastText' in order to obtain the features applicable in classification.

### 3.5. Classification
We experimented with five machine learning and two deep learning algorithms. Specifically, we used Logistics Regression (LR), Support Vector Machine (SVM), Decision Tree (DT), Random Forest (RF), and Multinomial Naive Bayes (MNB) for classic ML as well as simple Recurrent Neural Network (simpleRNN) and Long-Short-Term Memory netowrk (LSTM).

### 3.6. Performance evaluation metrics
Model evaluation is the process of validating the model performance on the test data. In this study, we used the following model performance evaluation metrics, namely, Precision, Recall, F1-score and Accuracy, which are calculated on the numbers of True Positives (TP), False Positives (FP), True Negatives (TN), and False Negatives (FN), according to the following formulas.

$$Accuracy = \frac{TP + TN}{TP + TN + FP + FN} \quad (1)$$

$$Precision = \frac{TP}{TP + FP} \quad (2)$$

$$Recall = \frac{TP}{TP + FN} \quad (3)$$

$$F1 - score = 2 \times \frac{Precision \times Recall}{Precision + Recall} \quad (4)$$

| Model | Vulgar | | | Non Vulgar | | | Accuracy |
|---|---|---|---|---|---|---|---|
| | Precision | Recall | F1-score | Precision | Recall | F1-score | |
| LR | 0.80 | 0.92 | 0.86 | 0.91 | 0.76 | 0.83 | 0.91 |
| SVM | 0.65 | 0.72 | 0.68 | 0.68 | 0.60 | 0.63 | 0.66 |
| DT | 0.62 | 0.86 | 0.72 | 0.77 | 0.47 | 0.58 | 0.67 |
| RF | 0.67 | 0.94 | 0.79 | 0.90 | 0.53 | 0.67 | 0.87 |
| MNB | 0.81 | 0.91 | 0.86 | 0.90 | 0.79 | 0.84 | 0.84 |

Table 4: Model performance using count vectorizer.

| Model | Vulgar | | | Non Vulgar | | | Accuracy |
|---|---|---|---|---|---|---|---|
| | Precision | Recall | F1-score | Precision | Recall | F1-score | |
| LR | 0.82 | 0.92 | 0.87 | 0.90 | 0.80 | 0.85 | 0.91 |
| SVM | 0.81 | 0.89 | 0.85 | 0.88 | 0.79 | 0.83 | 0.84 |
| DT | 0.56 | 0.96 | 0.71 | 0.85 | 0.21 | 0.34 | 0.67 |
| RF | 0.64 | 0.97 | 0.77 | 0.94 | 0.45 | 0.61 | 0.88 |
| MNB | 0.80 | 0.91 | 0.85 | 0.89 | 0.77 | 0.83 | 0.83 |

Table 5: Model performance using TF-IDF.

| Training | Testing |
|---|---|
| 80% | 20% |
| 1,988 | 497 |

Table 6: Training and test data ratio.

| Training | Validation | Testing |
|---|---|---|
| 70% | 15% | 15% |
| 1,741 | 372 | 372 |

Table 7: Training, validation and test data ratio.

## 4. Experimental results and discussion
### 4.1. Machine learning models for vulgarity detection

Below are the results of classic ML algorithms applied in vulgarity detection in Chittagonian language, specifically, LR, SVM, DT, RF and MNB using Count Vectorizer. Count Vectorizer is a process of converting textual features into numerical values, specifically into numbers representing basic frequency of words in the text. To train and test the models we divided the dataset in 80-20 ratio for training and testing set, respectively, as represented in Table 6.

The results achieved by ML models using this feature extraction technique were represented in Table 4.

Table 4 shows the overall performance of machine learning models using Count Vectorizer feature extraction technique. Here, LR model outperforms other models with the highest accuracy of 0.91. Class-wise performance also achieved higher precision and recall values for both classes. The second-best model was RF, which achieved 0.87 but in terms of precision, and recall MNB also performed well compared to LR.

On the contrary, TF-IDF stands for Term Frequency-Inverse Document Frequency of records. It assesses the word's relevance within a corpus or a dataset. The frequency of a term in the corpus represents how many times it appears in the text.

The highest score was also achieved by LR, with an accuracy of 0.91, and precision, recall and F1-score also scoring high for both classes. MNB achieved 0.83 of accuracy, less then RF but in terms of other metrics, the results were well balanced, as shown in Table 5.

### 4.2. Deep learning models for vulgarity detection

Deep Learning-based models like RNN, LSTM, or GRU have shown great achievements in various NLP tasks in recent years. In this study we used RNN and LSTM with Word2vec and fastText used for generating word embeddings. We divided the dataset into three parts, namely, train, validate (to check for overfitting) and test (to evaluate the model) as shown Table 7.

Using Word2vec word embeddings SimpleRNN outperformed the LSTM model achieving 0.84 of accuracy while LSTM achieved 0.63. In terms of precision and recall for both classes, SimpleRNN also provided better results, as represented in Table 8.

Also, both applied Deep Learning models showed better performance using fastText word embedding technique. Here also SimpleRNN acquired 0.90 of accuracy and more importantly models performed quite well in detecting both classes as shown in Table 8.

## 5. Conclusion and future work

In this study, we focused on the detection of vulgar remarks in social media posts using ML and DL classifiers, namely LR, SVM, DT, RF, MNB, simpleRNN, and LSTM with various feature extraction techniques such as Count Vectorizer, TF-IDF Vectorizer, Word2vec and fastText. We have constructed a dataset of 2,485 comments where vulgar and non-vulgar were evenly distributed. In our study, LR with Count Vector- izer, or TF-IDF Vectorizer, as well as simpleRNN with Word2vec and fastText were an effective approach for detecting vulgar remarks.

Based on the performed study, in the future, we plan to pursue a resource-constrained strategy for recognizing vulgarity, mostly focusing on the Chittagonian dialect. As in this study for the classification we used only the simplest baseline methods, we will apply other more robust methods, such as bidirectional RNNs (Schuster and Paliwal, 1997), and transformers (Aurpa et al., 2022).

| Word2vec | Vulgar | | | Non Vulgar | | | Accuracy |
|---|---|---|---|---|---|---|---|
| | Precision | Recall | F1-score | Precision | Recall | F1-score | |
| SimpleRNN | 0.78 | 0.98 | 0.86 | 0.97 | 0.70 | 0.81 | 0.84 |
| LSTM | 0.61 | 0.81 | 0.70 | 0.68 | 0.45 | 0.54 | 0.63 |
| fastText | Vulgar | | | Non Vulgar | | | Accuracy |
| | Precision | Recall | F1-score | Precision | Recall | F1-score | |
| SimpleRNN | 0.94 | 0.87 | 0.90 | 0.87 | 0.94 | 0.90 | 0.90 |
| LSTM | 0.63 | 0.89 | 0.74 | 0.79 | 0.45 | 0.57 | 0.68 |

Table 8: Model performance using Word2vec and fastText

## 6. References

Akhter, Shahin et al., 2018. Social media bullying detection using machine learning on bangla text. In 2018 10th International Conference on Electrical and Computer Engineering (ICECE). IEEE.

Andersson, Lars-Gunnar and Peter Trudgill, 1990. Bad language. Blackwell by arrangement with Penguin Books.

Aurpa, Tanjim Taharat, Rifat Sadik, and Md Shoaib Ahmed, 2022. Abusive bangla comments detection on facebook using transformer-based deep learning models. Social Network Analysis and Mining, 12(1):24.

Awal, Md Abdul, Md Shamimur Rahman, and Jakaria Rabbi, 2018. Detecting abusive comments in discussion threads using naïve bayes. In 2018 International Conference on Innovations in Science, Engineering and Technology (ICISET). IEEE.

Cachola, Isabel, Eric Holgate, Daniel Preoţiuc-Pietro, and Junyi Jessy Li, 2018. Expressively vulgar: The socio-dynamics of vulgarity and its effects on sentiment analysis in social media. In Proceedings of the 27th International Conference on Computational Linguistics.

Cohen, Jacob, 1960. A coefficient of agreement for nominal scales. Educational and psychological measurement, 20(1):37–46.

Das, Mithun, Somnath Banerjee, Punyajoy Saha, and Animesh Mukherjee, 2022. Hate speech and offensive language detection in bengali. arXiv preprint arXiv:2210.03479.

Emon, Estiak Ahmed, Shihab Rahman, Joti Banarjee, Amit Kumar Das, and Tanni Mittra, 2019. A deep learning approach to detect abusive bengali text. In 2019 7th International Conference on Smart Computing & Communications (ICSCC). IEEE.

Eshan, Shahnoor C and Mohammad S Hasan, 2017. An application of machine learning to detect abusive bengali text. In 2017 20th International conference of computer and information technology (ICCIT). IEEE.

Faisal Ahmed, Md, Zalish Mahmud, Zarin Tasnim Biash, Ahmed Ann Noor Ryen, Arman Hossain, and Faisal Bin Ashraf, 2021. Bangla text dataset and exploratory analysis for online harassment detection. arXiv e-prints:arXiv–2102.

Holgate, Eric, Isabel Cachola, Daniel Preoţiuc-Pietro, and Junyi Jessy Li, 2018. Why swear? analyzing and inferring the intentions of vulgar expressions. In Proceedings of the 2018 conference on empirical methods in natural language processing.

Hussain, Md Gulzar and Tamim Al Mahmud, 2019. A technique for perceiving abusive bangla comments. Green University of Bangladesh Journal of Science and Engineering:11–18.

Khan, Muhammad Moin, Khurram Shahzad, and Muhammad Kamran Malik, 2021. Hate speech detection in roman urdu. ACM Transactions on Asian and Low-Resource Language Information Processing (TALLIP), 20(1):1–19.

LewisM, P, 2009. Ethnologue: Languagesoftheworld, sixteenthedition.

Mahmud, Tanjim, Sudhakar Das, Michal Ptaszynski, Mohammad Shahadat Hossain, Karl Andersson, and Koushick Barua, 2022. Reason based machine learning approach to detect bangla abusive social media comments. In Intelligent Computing & Optimization: Proceedings of the 5th International Conference on Intelligent Computing and Optimization 2022 (ICO2022). Springer.

Masica, Colin P, 1993. The indo-aryan languages. Cambridge University Press.

Mehl, Matthias R, Simine Vazire, Nairán Ramírez-Esparza, Richard B Slatcher, and James W Pennebaker, 2007. Are women really more talkative than men? Science, 317(5834):82–82.

Pinker, Steven, 2007. The stuff of thought: Language as a window into human nature. Penguin.

Pradhan, Rahul, Ankur Chaturvedi, Aprna Tripathi, and Dilip Kumar Sharma, 2020. A review on offensive language detection. Advances in Data and Information Sciences: Proceedings of ICDIS 2019:433–439.

Sazzed, Salim, 2021. Identifying vulgarity in bengali social media textual content. PeerJ Computer Science, 7:e665.

Schuster, Mike and Kuldip K Paliwal, 1997. Bidirectional recurrent neural networks. IEEE transactions on Signal Processing, 45(11):2673–2681.

Wang, Na, 2013. An analysis of the pragmatic functions of "swearing" in interpersonal talk. En: Griffith Working Papers in Pragmatics and Intercultural Communication, 6:71–79.

Wang, Wenbo, Lu Chen, Krishnaprasad Thirunarayan, and Amit P Sheth, 2014. Cursing in english on twitter. In Proceedings of the 17th ACM conference on Computer supported cooperative work & social computing.